\documentclass[10pt,twocolumn,letterpaper]{article}

\usepackage{cvpr}              

\usepackage[accsupp]{axessibility}

\usepackage[dvipsnames]{xcolor}

\usepackage{amsmath}
\usepackage{amssymb}
\usepackage{booktabs}

\usepackage{multirow}
\usepackage{tabularx}
\usepackage{adjustbox}

\definecolor{cvprblue}{rgb}{0.21,0.49,0.74}
\usepackage[pagebackref,breaklinks,colorlinks,citecolor=cvprblue]{hyperref}

\def\paperID{5311} 
\def\confName{CVPR}
\def\confYear{2024}

\title{MaskINT: Video Editing via Interpolative Non-autoregressive Masked Transformers}

\author{Haoyu Ma\textsuperscript{1,2}\thanks{Work done during an internship at GenAI, Meta}, 
Shahin Mahdizadehaghdam\textsuperscript{2}, 
Bichen Wu\textsuperscript{2}, 
Zhipeng Fan\textsuperscript{2}, 
Yuchao Gu\textsuperscript{3},\\
Wenliang Zhao\textsuperscript{2}, 
Lior Shapira\textsuperscript{2}, 
Xiaohui Xie\textsuperscript{1}\\
\textsuperscript{1}University of California, Irvine, 
\textsuperscript{2}GenAI, Meta, 
\textsuperscript{3}National University of Singapore\\
Project Page: \href{https://maskint.github.io/}{https://maskint.github.io}
}

\begin{document}

\maketitle

\begin{figure*}[!tb]
    \centering
    \includegraphics[width=\linewidth]{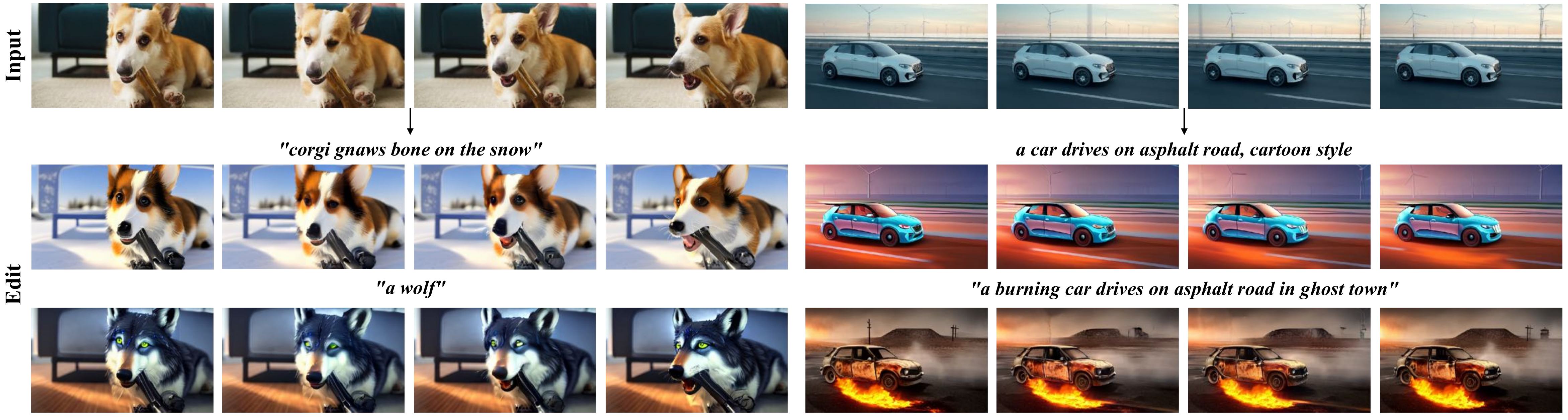}
    \vspace{-0.5em}
    \caption{ Examples of video editing with MaskINT.  }
    \label{fig:teaser}
\end{figure*}

\begin{abstract}
Recent advances in generative AI have significantly enhanced image and video editing, particularly in the context of text prompt control. State-of-the-art approaches predominantly rely on diffusion models to accomplish these tasks. However, the computational demands of diffusion-based methods are substantial, often necessitating large-scale paired datasets for training, and therefore challenging the deployment in real applications. To address these issues, this paper breaks down the text-based video editing task into two stages. First, we leverage an pre-trained text-to-image diffusion model to simultaneously edit few keyframes in an zero-shot way. Second, we introduce an efficient model called MaskINT, which is built on non-autoregressive masked generative transformers and specializes in frame interpolation between the edited keyframes, using the structural guidance from intermediate frames. Experimental results suggest that our MaskINT achieves comparable performance with diffusion-based methodologies, while significantly improve the inference time. This research offers a practical solution for text-based video editing and showcases the potential of non-autoregressive masked generative transformers in this domain.
\end{abstract}

\section{Introduction}
\label{sec:intro}

Text-based video editing, which aims to modify a video's content or style in accordance with a provided text description while preserving the motion and semantic layout, plays an important role in a wide range of applications, including advertisement, live streaming, and the movie industry, etc. This challenging task requires that edited video frames not only match the given text prompt but also ensure the consistency across all video frames. 
%


Recently, numerous studies have showcased the impressive capabilities of diffusion models \cite{ho2020denoising} in image generation and editing with text prompts \cite{rombach2022high,dai2023emu,peebles2023scalable,saharia2022photorealistic,zhang2023adding,brooks2023instructpix2pix,tumanyan2023plug}. 
When extending these foundation text-to-image (T2I) models to video generation and editing, a challenge is to maintain the temporal consistency. 
Existing solutions can be mainly divided into two ways: 
One is to finetune T2I models with additional temporal modules on paired text-video datasets \cite{ho2022imagen,esser2023structure,chai2023stablevideo,liew2023magicedit,blattmann2023stable}. However, the lack of extensive paired text-video datasets hinders these works from attaining the same level of performance as seen in image editing. 
The other involves leveraging a T2I models in a zero-shot manner \cite{geyer2023tokenflow,qi2023fatezero,khachatryan2023text2video,ceylan2023pix2video,zhang2023controlvideo}. 
These works find that extending the spatial self-attention across frames can produce consistent content across frames \cite{wu2022tune}. However, extended attention can lead to inconsistent editing of details due to its implicit temporal constraints. 
Moreover, while these techniques can generate high-fidelity videos, using cumbersome diffusion models to generate all frames is time-consuming. The integration of the extended spatial-temporal attention further exacerbates the running time, making them less practical for real-world applications. This raises the question: \textit{Is it necessary to utilize tedious diffusion models to generate all video frames? }

In this paper, we propose \textit{MaskINT}, an efficient prompt-based video editing framework designed to generate high-quality videos swiftly (\cref{fig:teaser}). 
MaskINT disentangles the task into two separate stages. 
In the first stage, we utilize pre-trained T2I models with extended attention to jointly edit only two keyframes (i.e., the initial and last frames) from the video clip, guided by the provided text prompt. 
In the second stage, we introduce a novel structure-aware frame interpolation module based on non-autoregressive generative Transformers \cite{chang2022maskgit,chang2023muse,yu2023magvit}, which generates all intermediate frames in parallel with structural cues and iteratively refine predictions in a few steps. 
Our framework offers several advantages: 
First, through the disentanglement of frame editing and interpolation, MaskINT eliminates the requirement for paired text-video datasets during training, thereby enabling us to train the MaskINT using large-scale video-only datasets. 
Second, the usage of non-autoregressive generation significantly accelerates the processing time.  
Experimental results indicate that MaskINT achieves comparable performance with pure diffusion methods in terms of temporal consistency and alignment with text prompts, while provides 5-7 times faster inference time. 
Our work not only provides a practical solution for text-based video editing in terms of the trade-off between quality and efficiency, but also highlights the potential of masked generative transformers in this domain.


Our major contributions are summarized as follows:
\begin{itemize}
    \item We introduce MaskINT, a two-stage pipeline for text-based video editing, which involves keyframes joint editing and structure-aware frame interpolation, eliminating the need for paired text-video datasets. 
    
    \item The proposed structure-aware frame interpolation is the pioneer work that explicitly introduces structure control into non-autoregressive generative transformers. 
    
    \item Experimental results demonstrate that MaskINT achieves comparable performance with diffusion methods, while significantly reducing the inference time. 
    
\end{itemize}

\section{Related Work}

\paragraph{Generative Transformers. }
Following GPT \cite{brown2020language}, previous works \cite{esser2021taming, yu2022vectorquantized,le2021ccvs, ge2022long,hong2022cogvideo} first tokenize images/videos into discrete tokens, and train \textit{Autoregressive Generative Transformers} to perform image/video generation, where tokens are generated sequentially based on previous outputs. 
However, these methods become exceedingly time-consuming when the length of the token sequence increases.

Recently, \textit{Non-autoregressive Generative Transformers} have emerged as efficient solutions \cite{chang2022maskgit,chang2023muse,yu2023magvit,gupta2023maskvit,villegas2023phenaki}. 
These works train bidirectional transformers \cite{devlin2018bert} with masked token modeling (MTM). During inference, all discrete tokens are generated in parallel and refined iteratively in a few steps, known as non-autoregressive decoding. 
Specifically, in image generation, MaskGiT \cite{chang2022maskgit} first shows the capability and efficiency  of non-autoregressive transformers. It can be seamlessly extended to tasks like inpainting and outpainting by applying various initial mask constraints. 
Muse \cite{chang2023muse} achieves state-of-the-art performance in text-to-image generation by training on large-scale text-image datasets and brings significantly efficiency improvement. 
StyleDrop \cite{sohn2023styledrop} further finetunes Muse with human feedback to perform text-to-image generation guided with a reference style image. 
MaskSketch \cite{bashkirova2023masksketch} introduces implicit structural guidance into MaskGiT by calculating the similarity of attention maps in the sampling step. 
In video generation, 
MaskViT \cite{gupta2023maskvit} employ 2D tokenizer and trains a bidirectional window transformer to perform frame prediction. 
Phenaki \cite{villegas2023phenaki} trains a masked transformer to generate short video clips condition on text prompt and extends it to arbitrary long video with different prompts in an autoregressive way. 
MAGVIT \cite{yu2023magvit} utilizes 3D  tokenizer to quantize videos and trains a single model to perform multiple video generation tasks such as inpainting, outpainting, frame interpolation, etc. 
However, to the best of our knowledge, there is currently no existing literature in the field of text-based video editing utilizing masked generative transformers. Besides, there is a notable absence of research that delves into explicit structural control within this area.

\vspace{-0.5em}
\paragraph{Diffusion Models in Image Editing. }
Built upon the latent diffusion models (LDM) \cite{rombach2022high}, numerous studies have achieved significant success in text-based image editing \cite{zhang2023adding, mou2023t2i,tumanyan2023plug,kawar2023imagic,hertz2022prompt,zhang2023sine,parmar2023zero, meng2021sdedit, couairon2022diffedit}. 
For example, ControlNet \cite{zhang2023adding}, T2I-Adapter \cite{mou2023t2i}, and Composer \cite{huang2023composer} finetune LDM with spatial condition such as depth maps and edge maps, enabling text-to-image synthesis with the same structure as the input image.  
The PNP \cite{tumanyan2023plug} incorporates DDIM inversion features \cite{song2020denoising} from the input image into the T2I generation process, enabling zero-shot image editing with pre-trained T2I models. 
InstructPix2Pix \cite{brooks2023instructpix2pix} trains a conditional diffusion model for text-guided image editing using synthetic paired examples, which avoid the tedious inversion. 
Nevertheless, applying these T2I models on each frame independently usually leads to inconsistencies and flickering.

\vspace{-0.5em}
\paragraph{Diffusion Models in Video Editing. }
Diffusion models also dominate the field of video generation and video editing \cite{singer2022make,ho2022imagen,chai2023stablevideo,wu2022tune}. 
Due to the lack of billion-scale paired video datasets, these works usually finetune existing T2I foundation models \cite{rombach2022high} with additional temporal modules on paired text-video datasets \cite{esser2023structure} or overfit a single video \cite{wu2022tune}. 
Meanwhile, several works explore using T2I models in a zero-shot manner for video editing \cite{wu2022tune,yang2023rerender,qi2023fatezero,khachatryan2023text2video,zhang2023controlvideo,wang2023videocomposer,ceylan2023pix2video}. To enable a cohesive appearance, a common approach in these studies involves extending the spatial self-attention module to spatial-temporal attention module and conducting cross-frame attention. 
In detail, 
Text2Video-Zero \cite{khachatryan2023text2video} performs cross-frame attention of each frame on the first frame to preserve appearance consistency. 
ControlVideo \cite{zhang2023controlvideo} extends ControlNet \cite{zhang2023adding} with fully cross-frame attention to joint edit all frames and further improves the performance with interleaved-frame smoother. 
TokenFlow \cite{geyer2023tokenflow} enhance PNP \cite{tumanyan2023plug} with extended-attention to jointly edit a few keyframes at each denoising step and propagate the edit frames to the entire video. 
However, conducting hundreds of denoising steps with extended attention across all frames is time-consuming. 
In contrast, we only utilize diffusion models to edit a few keyframes, rather than all frames. 


\vspace{-0.5em}
\paragraph{Video Frame Interpolation (VFI).}
VFI aims to generate intermediate images between a pair of frames, which can be applied to creating slow-motion videos and enhancing refresh rate. Advanced methods typically estimate dense motions between frames, like optical flow, and subsequently warping the provided frames to generate intermediate ones \cite{niklaus2017video,jiang2018super,sim2021xvfi,lu2022video,reda2022film,huang2022real}. However, these regression methods are most effective with simple or monotonous motion. 
In our work, we design a generative model to perform frame interpolation with additional structural signals, enabling interpolation with large motion.

\section{Preliminaries}
\subsection{Masked Generative Transformers}
\vspace{-0.3em}
Masked generative transformers \cite{chang2022maskgit} follow a two-stage pipeline. In the first stage, an image is quantized into a sequence of discrete tokens via a Vector-Quantized (VQ) auto-encoder \cite{esser2021taming}. 
In detail, given an image $\textbf{I} \in \mathbb{R}^{H \times W \times 3}$, an encoder $\mathcal{E}$ encodes it into a series of latent vectors, which are then discretized via nearest neighbor lookup in a codebook of quantized embeddings with size $M$. 
Thus, an image can be represented with a sequence of codebook's indices $\mathbf{Z} = [z_i]_{i=1}^{h \times w}, z_i \in \{1,2,...,M\}$, where $h$ and $w$ is the resolution of latent features. A decoder $\mathcal{D}$ can reconstruct the indices back to image $\mathcal{D}(\mathbf{Z}) \approx \mathbf{I}$. 
In the second stage, a bidirectional transformer model \cite{vaswani2017attention} is learned with Masked Token Modeling (MTM). 
Specifically, during training, a random mask ratio $r \in (0,1)$ is selected and $\lfloor\gamma(r) \cdot h \times w \rfloor$ tokens in $Z$ are replaced with a special $\texttt{[MASK]}$ token, where $\gamma(r)$ is a mask scheduling function \cite{chang2022maskgit}. We denote the corrupted sequence with masked tokens as $\mathbf{\Bar{Z}}$ and conditions such as class labels or text prompt as $\mathbf{c}$. 
Given the training dataset $\mathbb{D}$, a BERT \cite{devlin2018bert} parameterized by $\Phi$ is learned to minimize the cross-entropy loss between the predicted and the ground truth token at each masked position: 
\begin{equation}
    \mathcal{L}_{MTM} = \mathop{\mathbb{E}}_{\mathbf{Z} \in \mathbb{D} } \left[ \sum_{ \Bar{z_i}= \texttt{[MASK]} }  -\log p_{\Phi}(z_i | \mathbf{\Bar{Z}}, \mathbf{c}) \right]. 
\end{equation}
During inference, non-autoregressive decoding is applied to generate images. Specifically, all tokens of an image or video are initialized as \texttt{[MASK]} tokens. At step $k$, all tokens are predicted in parallel while only tokens with the highest prediction scores are kept. The remaining tokens with least prediction scores are masked out and regenerated in the next iteration. The mask ratio is determined by $\gamma(\frac{k}{K})$ at step $k$, where $K$ is the total number of iteration steps.

\subsection{ControlNet}
\vspace{-0.3em}
\paragraph{Latent Diffusion Models} 
Denoising Diffusion Probabilistic Models \cite{ho2020denoising} generate images through a progressive noise removal process applied to an initial Gaussian noisy image, carried out over a span of $T$ time steps.
To enable efficient high-resolution image generation, LDM \cite{rombach2022high}  operates the diffusion process in the latent space of an autoencoder. In detail, an encoder $\mathcal{E'}$ first compresses an image $\mathbf{I}$ to a low-resolution latent code $x=\mathcal{E'}(\mathbf{I}) \in \mathbb{R}^{h \times w \times c}$. Second, a U-Net $\epsilon_\theta$ with attention modules \cite{vaswani2017attention} is trained to remove the noise with loss  
\begin{equation}
    \mathcal{L}_{LDM} =  \mathbb{E}_{x_0, \epsilon \sim N(0, I), t\sim  } \| \epsilon - \epsilon_\theta (x_t, t, \tau) \|_2^2
\end{equation}
, where $\tau$ is the text prompt and $x_t$ is the noisy latent sample at timestep $t$. 

\paragraph{ControlNet} 
Since accurately describing the layout of generated images using only text prompts is challenging, ControlNet \cite{zhang2023adding}  extends LDM by incorporating spatial layout conditions such as edge maps, depth maps, and human poses. In detail, ControlNet train the same U-Net architecture as LDM and finetune it with additional conditions. We denote it as $\epsilon_\theta'(x_t, t, \tau, s)$, where $s$ is the spatial layout.

\begin{figure*}[!ht]
    \centering
    \includegraphics[width=0.9\linewidth]{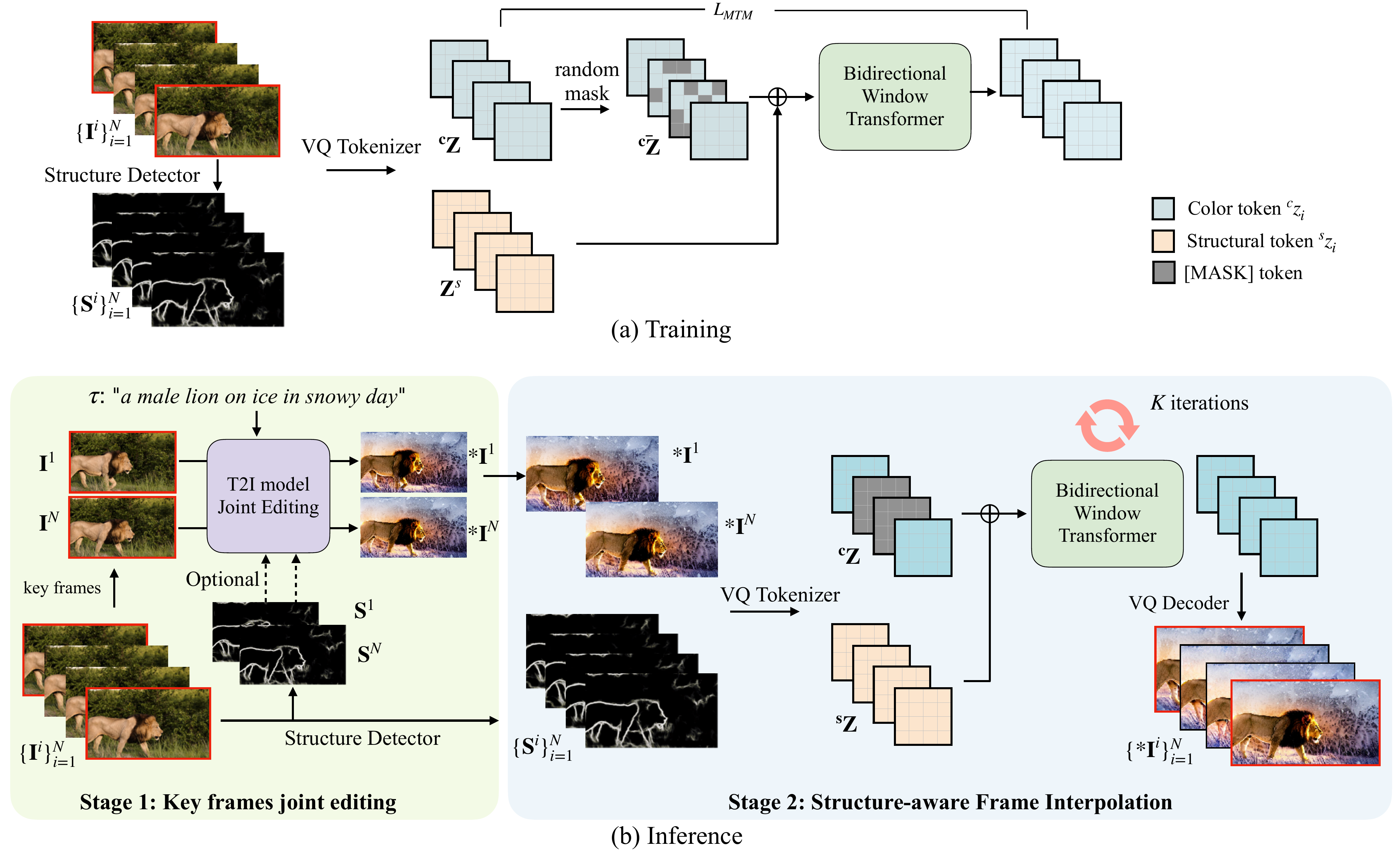}
    \vspace{-0.5em}
    \caption{Overview of MaskINT. MaskINT disentangle the video editing task into two separate stages, i.e., keyframes joint editing and structure-aware frame interpolation. }
    \label{fig:framework}
\end{figure*}

\section{Methodology}

\vspace{-0.3em}
\subsection{Overview}
\vspace{-0.3em}

\cref{fig:framework} shows an overview of our MaskINT. 
MaskINT disentangle the video editing task into keyframe editing stage and structure-aware frame interpolation stage. 
Specifically, given a video clip with $N$ frames $ \{\mathbf{I}^i\}_{i=1}^N$, 
in the first stage, with the input text prompt $\mathbf{\tau}$, we simultaneously edit two keyframes, i.e., the initial frame $\mathbf{I}^1$ and last frame $\mathbf{I}^N$, using existing image-editing model $g(\cdot)$ in a zero-shot way.  
This process results in the creation of high-quality coherent edited frames $\mathbf{^*I}^1, \mathbf{^*I}^N = g(\mathbf{I}^1, \mathbf{I}^N, \mathbf{\tau})$. 
In the second stage, MaskINT performs structure-aware frame interpolation via non-autoregressive transformers. The entire edited video frames are generated by $\{ \mathbf{^*I}^i \}_{i=1}^{N} = f_\Phi(\mathbf{^*I}^1, \mathbf{^*I}^N,  \{ \mathbf{S}^i\}_{i=1}^N )$,  where $\mathbf{S}^i \in [0,1]^{H\times W \times 1}$ is the structural condition (i.e., the HED edge map \cite{xie2015holistically}).   
The structure-aware interpolation module is trained with masked token modeling (MTM) on video only datasets, conditioning on the structural signal $ \{ \mathbf{S}^i\}_{i=1}^N $ as well as  the initial frame $\mathbf{I}^1$ and last frame $\mathbf{I}^N$.

\vspace{-0.3em}
\subsection{Keyframes Joint Editing}
\vspace{-0.5em}

To maintain the structure layout, we take ControlNet to edit $\mathbf{I}^1$ and $\mathbf{I}^N$ based on text prompt $\tau$ as well as their edge maps $\mathbf{S}^1$ and $\mathbf{S}^N$. 
However, even with the identical noise, applying ControlNet to each keyframes individually cannot guarantee the appearance consistency. 
To address this issue, following previous work \cite{wu2022tune, geyer2023tokenflow, zhang2023controlvideo}, we extend the self-attention to simultaneously process two keyframes. 
In detail, the self-attention module projects the noisy feature map $x_t^j$ of $j^{th}$ frame at time step $t$ into query $\mathbf{Q}^j$, key $\mathbf{K}^j$, value  $\mathbf{V}^j$ in the original U-Net. 
We extend the self-attention block to perform attention across all selected keyframes by concatenating their keys and values: 
\begin{equation}
    \label{eq:joint_attn}
     \text{Softmax}(\dfrac{ \mathbf{Q}^j [\mathbf{K}^1, \mathbf{K}^N ]^\mathbf{T} }{ \sqrt{c} } ) [\mathbf{V}^1, \mathbf{V}^N].
\end{equation}
Since each frame queries all frames and aggregates information from all of them, the appearance consistency can be achieved. Although only two frames were used in \cref{eq:joint_attn}, this joint editing can be seamlessly extended to any number of frames, but at the cost of increased processing time and significant GPU memory demand.

\subsection{Structure-Aware Frame Interpolation}
\vspace{-0.3em}

\paragraph{Structural-aware embeddings. }
Non-autoregressive masked generative transformers \cite{gupta2023maskvit,yu2023magvit}  can effectively perform frame prediction and interpolation tasks. However, these works lack explicit control of structure, making it difficult to follow the motion of input videos. 
Thus, we explicitly introduce structural condition of each frame into the generation process. 
In detail, we tokenize both RGB frames $ \{\mathbf{I}^i\}_{i=1}^N$ and structure maps  $ \{\mathbf{S}^i\}_{i=1}^N$ with an off-the-shelf 2D VQ tokenizer \cite{esser2021taming}. 
We utilize 2D VQ rather than 3D VQ \cite{yu2023magvit} to accommodate diverse numbers of frames and frame rate. 
We denote tokens from RGB frames as $\mathbf{^cZ} = \{ ^cz_i\}_{i=1}^{N \times h \times w}$ (color token) and tokens from edge maps as $\mathbf{^sZ}=\{ ^sz_i\}_{i=1}^{N \times h \times w}$ (structure token), where $^sz_i$, $^cz_i \in \{1,2,...,M\}$ and $M$ is the codebook size. 
Subsequently, two distinct embedding layers $e^C(\cdot)$ and $e^S(\cdot)$ are learned to map tokens $\mathbf{^cZ}$ and $\mathbf{^sZ}$ into their respective embedding spaces. Learnable 2D spatial positional encoding $\mathbf{P^S}$ and temporal positional encoding $\mathbf{P^T}$ are also added \cite{bertasius2021space}. Thus, the input to the following transformer layers can be formulated by $\mathbf{X} = e^c(\mathbf{^cZ}) + e^s(\mathbf{^sZ}) + \mathbf{P^S} + \mathbf{P^T} \in \mathbb{R}^{N \times h \times w \times c}$. 


%
%
\vspace{-0.5em}
\paragraph{Transformer with Window-Restricted Attention. }
Previous masked generative transformers \cite{chang2022maskgit,yu2023magvit} employ the standard transformer with global attention \cite{vaswani2017attention}. 
Given that there is no substantial motion between consecutive frames, we adopt self-attention within a restricted window \cite{gupta2023maskvit} to further mitigate computational overhead
In detail, our approach involves two distinct stages of attention. First, we conduct spatial window attention, where attention is calculated within each frame of dimensions $1 \times h \times w$. We then conduct spatial-temporal window attention, where attention is calculated within a tube of dimensions $N \times h_w \times w_w$, where $h_w$ and $w_w$ is the window size. 
Besides, to further reduce computation of transformer, we also add a shallow convolution layers to downsample $\mathbf{X}$ before the transformer encoder layers and an upsample layer at the end.

\vspace{-0.5em}
\paragraph{Training. }
We train MaskINT using video-only datasets. 
Denote the color token of $i^{th}$ RGB frame as $\mathbf{^cZ}^i=\{^cz^i\}_{i=1}^{h\times w}$. 
During the training time, we keep color tokens of the initial frame $\mathbf{^cZ}^1$ and last frame $\mathbf{^cZ}^N$, and randomly replace $[\gamma(r) \cdot (N-2)\cdot N] $  color tokens of intermediate frames with the \texttt{[MASK]} tokens. We denote this corrupted video color tokens as $\mathbf{\Bar{^cZ}} = \{ \mathbf{^cZ}^1, \mathbf{\Bar{^cZ}}^2, ..., \mathbf{\Bar{^cZ}}^{N-1}, \mathbf{^cZ}^N \}$. 
The structure-aware window-restricted transformer with parameters $\Theta$ is trained by 

\begin{equation}
    \mathcal{L}_{MTM} = \mathop{\mathbb{E}}_{ \{ \mathbf{^cZ}, \mathbf{^sZ} \} \in 
\mathbb{D} } [ \sum_{ \Bar{^cz_i}= \texttt{[MASK]} }  -\log p_{\Theta}(^cz_i | \mathbf{\Bar{^cZ}}, \mathbf{^sZ})  ].
\end{equation}

\vspace{-0.5em} 
\paragraph{Inference. }
During inference, MaskINT can seamlessly generalize to perform frame interpolation between the jointly edited frames, although it is only trained with regular videos. 
Specifically, we tokenize the the initial and last edited frames $\mathbf{^*I}^1$ and $\mathbf{^*I}^N$ from Stage 1 into color tokens $\mathbf{^{c}_*Z}^1$ and $\mathbf{^{c}_*Z}^N$, and initialize color tokens of all intermediate frames $\{ \mathbf{\Bar{^cZ}}^2, ..., \mathbf{\Bar{^cZ}}^{N-1} \}$ with \texttt{[MASK]} tokens. 
We follow the iterative decoding in MaskGiT \cite{chang2022maskgit} with a total number of $K$ steps. At step $k$, we predict all color tokens in parallel and keep tokens with the highest confidence score.

\vspace{-0.3em}
\section{Experiments}
\vspace{-0.3em}

\subsection{Settings}
\vspace{-0.3em}
\paragraph{Implementation Details. }
We train our model with $100k$ videos from ShutterStock \cite{SSdataset}. During training, we random select a  video clip with $T=16$ frames and frame interval of $1,2,4$ from each video. All frames are resized to $384\times 672$. We utilize Retina-VQ \cite{dubeyretinavq} with $8$ downsample ratio, i.e., each frame has $48\times84$ tokens. 
We employ Transformer-Base as our MaskINT and optimized it from scratch with the AdamW optimizer \cite{loshchilov2018decoupled} for a duration of $100$ epochs. The initial learning rate is set to $1e-4$ and decayed with cosine schedule. 
We set the number of decoding step $K$ to $32$ and the temperature $t$ to $4.5$ in inference.

\vspace{-0.5em} 
\paragraph{Evaluation. } %
We evaluate our method on $40$ selected object-centric videos \cite{wu2022tune} from the DAVIS dataset \cite{Pont-Tuset_arXiv_2017}, as well as $30$ unseen videos from the ShutterStock \cite{SSdataset}, covering animals, vehicles, etc. For each video, we manually design $5$ edited prompts, including object editing, background changes and style transfers. 
Following previous works \cite{zhang2023controlvideo,qi2023fatezero,geyer2023tokenflow,ceylan2023pix2video}, we evaluate the quality of the generated videos with the following metrics: 
1) Prompt consistency, which calculates the CLIP \cite{radford2021learning} embedding similarity between text prompt and all video frames. 
2) Temporal consistency, which calculates the CLIP embedding similarity of all pairs of consecutive frames. 
3) Warpping-error \cite{lai2018learning}, which computes the optical flow between consecutive frames using RAFT \cite{teed2020raft}, and warps the edited frames to the next according it. 
4) Long-term temporal consistency. Inspired by the global coherence metric in \cite{tzaban2022stitch}, we also calculate the warp error between all possible pairs of frames as the long-term consistency. 
As for efficiency, we report the running time for generating a 16-frame video clip on a single NVIDIA A6000 GPU.

\begin{figure*}[!ht]
    \centering
    \includegraphics[width=\linewidth]{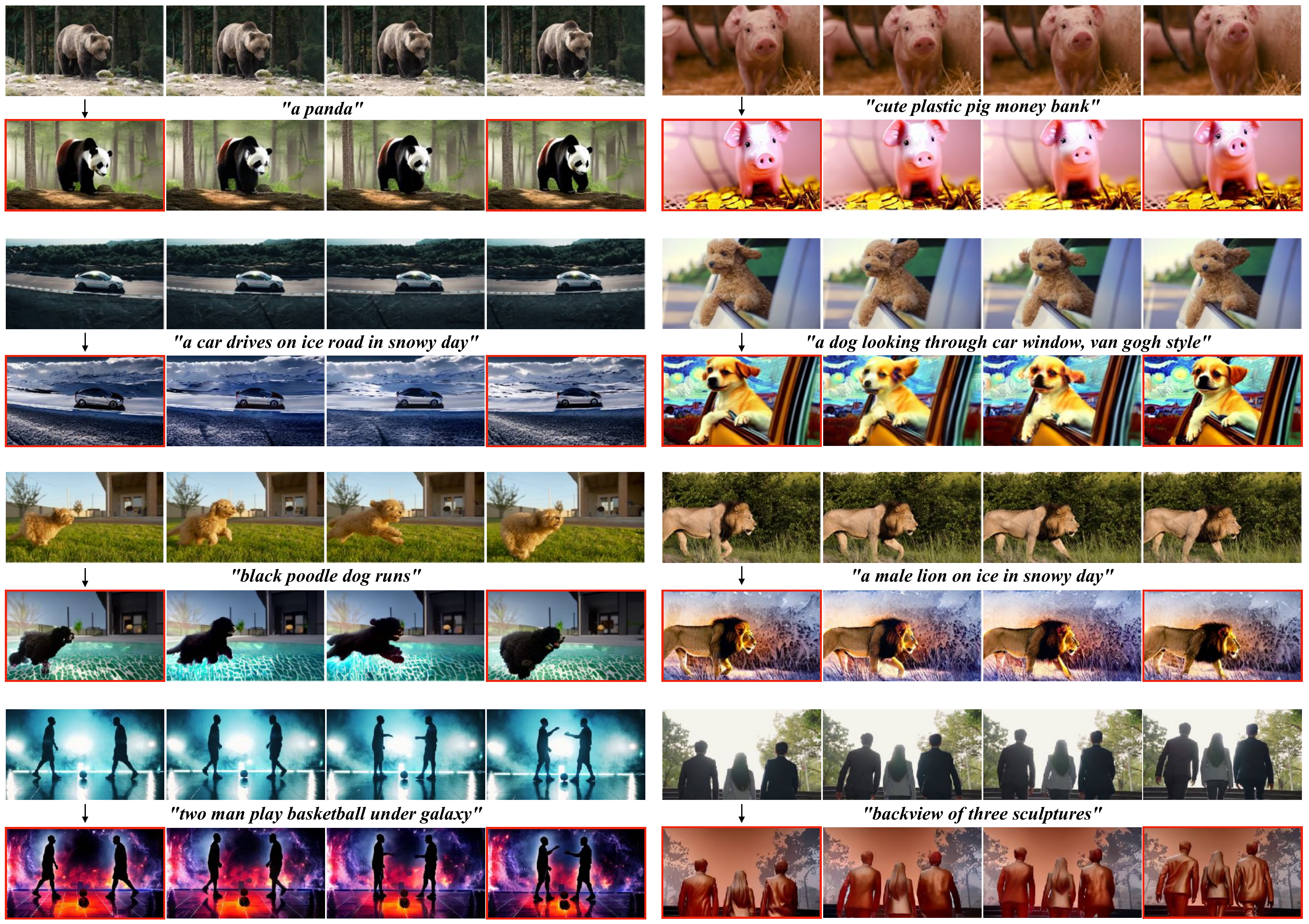}
    \caption{ \small{Examples of video editing with MaskINT. Frames with red bounding box are jointly edited keyeframes. } }
    \label{fig:example}
\end{figure*}

\vspace{-0.3em} 
\subsection{Results}
\vspace{-0.3em} 
We apply ControlNet to each frame individually  with the same noise as the baseline. 
We select methods that built upon T2I diffusion models for comparison, including Tune-A-Video \cite{wu2022tune}, TokenFlow \cite{geyer2023tokenflow}, Text-to-video zero \cite{khachatryan2023text2video}, and ControlVideo \cite{zhang2023controlvideo}. 
Besides, we compare with VFI methods such as FILM \cite{reda2022film} using the same edited keyframes. 


\begin{table}[!ht]
    \centering
    \resizebox{1.0\linewidth}{!}{
    \begin{tabular}{l|cccc|cccc|c}
        \toprule
        \multirow{3}{*}{Method} & \multicolumn{4}{c|}{DAVIS} & \multicolumn{4}{c|}{ShutterStock} & \multirow{3}{*}{Time} \\
         & P.C.$\uparrow$ &  T.C.$\uparrow$  & W.E.$\downarrow$ & L.C.$\downarrow$ & P.C$\uparrow$ &  T.C.$\uparrow$  & W.E$\downarrow$ & L.C.$\downarrow$ \\
         & & & /$\times10^{-3}$ & /$\times10^{-3}$ & & & /$\times10^{-3}$ & /$\times10^{-3}$ \\


        \midrule 
        ControlNet per frame \cite{zhang2023adding}       & 0.314 & 0.914  & 36.6 & 58.6    & 0.304 & 0.942  & 25.9 & 45.1 & 50s \\
        Tune-a-Video \cite{wu2022tune}                    & 0.299 & 0.966  & 20.4 & 48.2    & 0.292 & 0.979  & 13.4 & 34.9 & 20min \\
        Text2Video-zero \cite{khachatryan2023text2video}  & 0.312 & 0.964  & 20.7 & 42.5    & 0.304 & 0.981  & 16.0 & 33.0 & 60s \\
        ControlVideo-edge \cite{zhang2023controlvideo}    & 0.314 & 0.975  & 6.9 & 23.9     & 0.303 & 0.986  & 7.4 & 20.5 & 120s \\
        TokenFlow \cite{geyer2023tokenflow}               & 0.317 & 0.977  & 7.0 & 19.6     & 0.313 & 0.987  & 5.4 & 15.8 &  150s \\
        MaskINT (ours)                                    & 0.311 & 0.952  & 9.5 & 27.7     & 0.304 & 0.971  & 8.6 & 22.3 & 22s \\
    \bottomrule
    \end{tabular}
    }
    \vspace{-0.5em}
    \caption{\footnotesize{Quantitative comparisons. ``T.C." stands for ``temporal consistency", ``P.C." stands for ``prompt consistency", ``W.E" stands for ``warpping-error", and ``L.C." stands for ``long-term temporal consistency". } }
    \label{tab:comparison}

    \vspace{-1.7em}
\end{table}

\vspace{-0.5em}
\paragraph{Quantitative Comparisons. }
\cref{tab:comparison} summarize the performance of these methods on both DAVIS and ShutterStock datasets. 
Notably, our method achieves comparable performance with diffusion methods, in terms of all four evaluation metrics, while brings a significant acceleration in processing speed. 
In detail, the test-time finetuning of TAV \cite{wu2022tune} is extremely time-consuming, making it a less practical solution. 
MaskINT is almost 5.5 times faster than ControlVideo \cite{zhang2023controlvideo}, whose fully cross-frame attention is computationally extensive. Moreover, MaskINT is nearly 7 times faster than TokenFlow \cite{geyer2023tokenflow}, whose DDIM inversion is time-consuming. 
The efficiency of our method is derived from a combination of a lightweight network and a reduced number of decoding steps with masked generative transformers.

\begin{figure*}[!ht]
    \centering
    \includegraphics[width=0.99\linewidth]{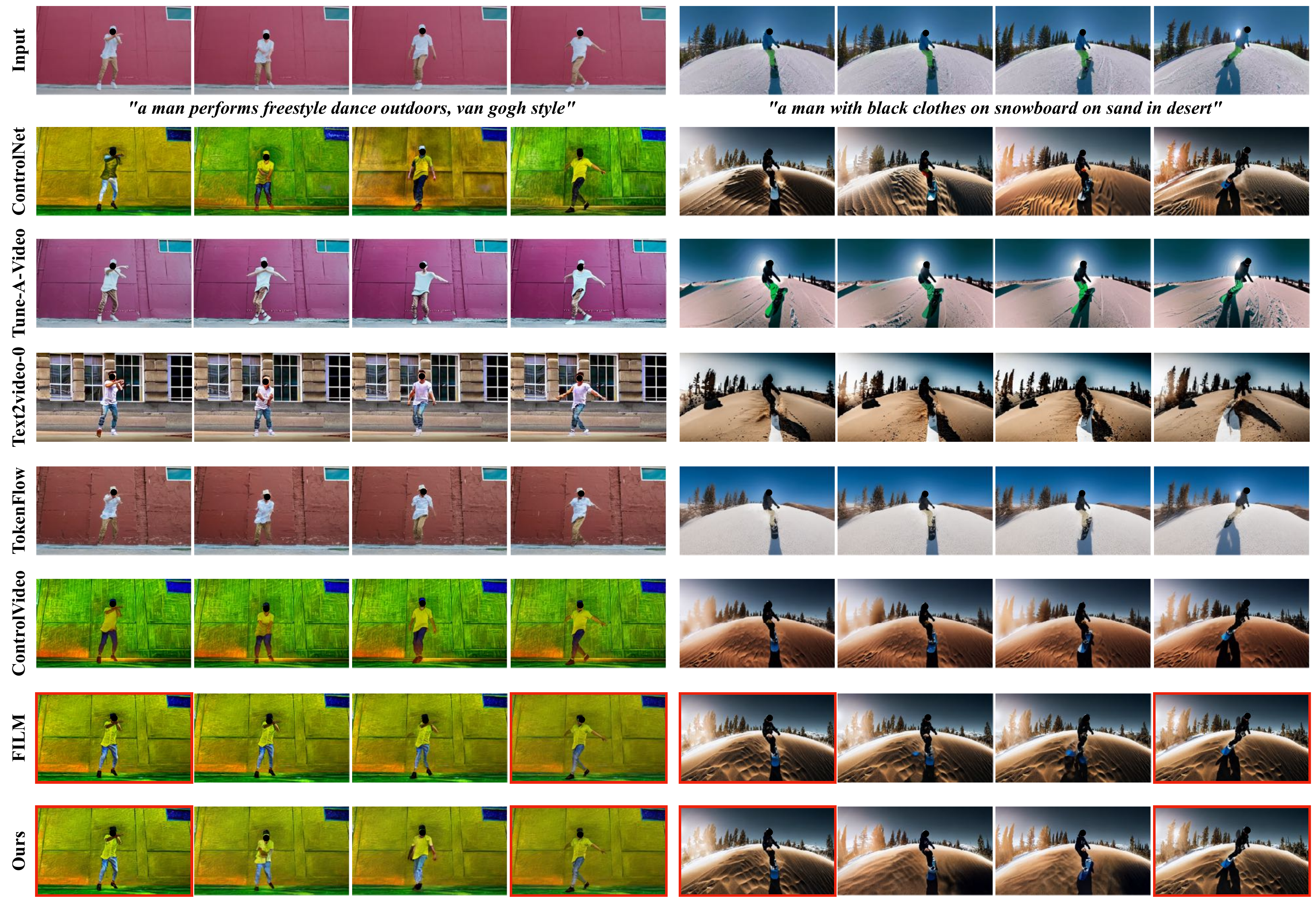}
    \caption{ \small{Qualitative comparisons with diffusion-based methods. More examples are shown in Supplementary. } }
    \label{fig:cmp}
\end{figure*}

\vspace{-0.5em}
\paragraph{Qualitative Comparisons. } 

\cref{fig:teaser} and \cref{fig:example} show edited video samples using MaskINT. 
Our method is able to generate temporally consistent videos in accordance with text prompts, enabling diverse applications including stylization, background and foreground editing, and beyond.
It is also effective in videos with complex motion, such as jumping and running, as well as scenarios with multiple subjects. 
Besides, although the structure-aware interpolation module is trained with realistic videos only, it still generalizes to diverse synthesized images. For example, we can successfully generate a video in the style of Van Gogh even in the absence of videos for this particular style in the training set. 

\cref{fig:cmp} provides a comparison of MaskINT to other methods. 
Remarkably, diffusion-based methods with extended attention  \cite{wu2022tune,geyer2023tokenflow,khachatryan2023text2video,zhang2023controlvideo} can maintain the overall appearance consistency, but sometimes result in inconsistent details. For example, TokenFlow \cite{geyer2023tokenflow} and Text2Video-Zero\cite{khachatryan2023text2video} exhibit noticeable artifacts in the leg region of the human subjects and ControlVideo \cite{zhang2023controlvideo} produces inconsistent hats. The potential explanation lies in the fact that these methods offer control over temporal consistency implicitly. 
Besides, FILM \cite{reda2022film} generates videos that cannot follow the original motions. 
Our MaskINT consistently interpolates the intermediate frames based on the structure condition and even maintain better consistency in local regions.

\begin{figure}[!ht]
    \centering
    \includegraphics[width=1.0\linewidth]{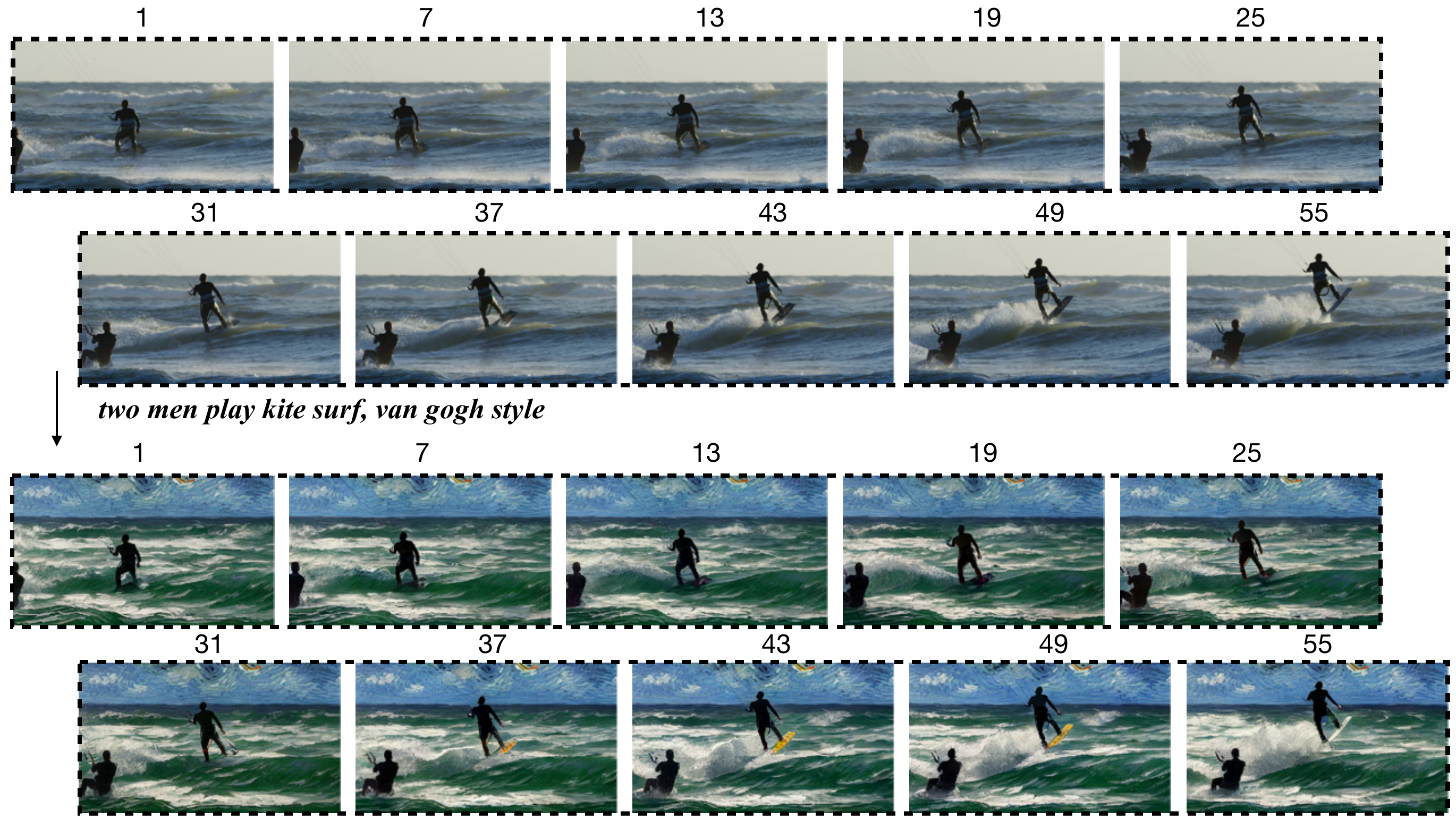}
    \vspace{-0.5em}
    \caption{ \small{Examples of long video generation. The number indicates the index of frame. More examples are in Supplementary. } }
    \label{fig:long_video}
\end{figure}

\vspace{-0.5em}
\paragraph{Extension on Long Video Editing. }
Given that  the non-autoregressive pipeline generates all frames simultaneously, it's challenging to edit an entire long video at once due to GPU memory limitation. 
Nevertheless, our framework can be extended to handle long videos by dividing them into short clips and progressively performing frame interpolation within each segment. 
For instance, given a video with $60$ frames, we select the $1^\text{st}$, $16^\text{th}$, $31^\text{st}$, $46^\text{th}$, and $60^\text{th}$ frames as keyframes. We jointly edit these selected $5$ frames together and then perform structure-aware frame interpolation within each pair of consecutive keyframes. 
As shown in \cref{fig:long_video}, with this design, our method can still generate consistent long videos. 
Moreover, in this proposed extension, the generation of later frames is dissociated from that of early frames, which differs from the autoregressive generation pipeline in Phenaki \cite{villegas2023phenaki}. 
Consequently, even if some early frames encounter difficulties, the generation of later frames can still proceed successfully.

\subsection{Ablation Studies}
\vspace{-0.3em}

\paragraph{Video Frame Interpolations.}

In this ablation study, we evaluate the performance of the structure-aware interpolation module. 
In detail, we perform frame interpolation using original keyframes, and compare the interpolated frames with the original intermediate video frames. 
We use same testing samples from DAVIS and Shutterstock datasets, employing peak signal-to-noise ratio (PSNR), learned perceptual image patch similarity (LPIPS), and structured similarity (SSIM) as the evaluation metrics. We benchmark our method against two state-of-the-art VFI methods, namely FILM \cite{reda2022film} and RIFE \cite{huang2022real}. We also showcase the performance of applying VQ-GAN \cite{esser2021taming} to all video frames, serving as an upper bound for our method. 

\begin{figure}[!ht]
    \centering
    \includegraphics[width=1.0\linewidth]{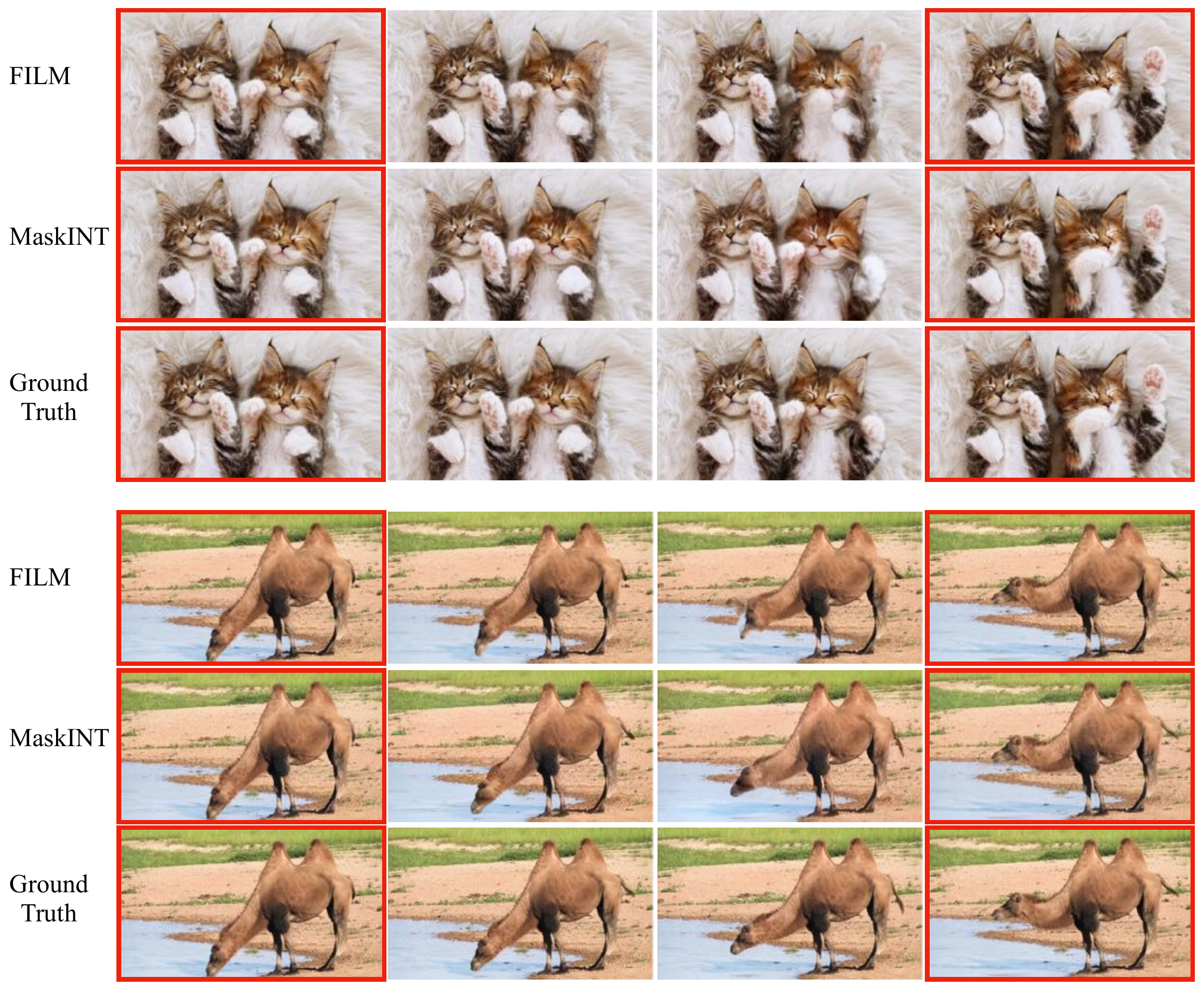}
    \caption{ \small{Qualitative comparisons on video reconstruction with original RGB frames. Frames with red bounding box are given. } }
    \label{fig:cmp_vfi}
\end{figure}

As in \cref{tab:rgb_recon}, our method significantly outperforms VFI methods on all evaluation metrics, with the benefit of the structure guidance from the intermediate frames. Furthermore, \cref{fig:cmp_vfi} shows qualitative comparison between FILM \cite{reda2022film} and MaskINT. Even when confronted with significant motion between two frames, our method successfully reconstructs the original video, maintaining consistent motion through the aid of structural guidance. In contrast, FILM introduces undesirable artifacts, including disorted background, duplicated cat hands, and the absence of a camel's head, etc. 
The major reason is that current VFI models mainly focus on generating slow-motion effects and enhancing frame rate, making them less effective in handling frames with complex motions, which usually requires a better semantic understanding. Besides, the absence of structural guidance poses a challenge for them in accurately aligning generated videos with the original motion.

\begin{table}[ht]

    \centering
    \resizebox{1.0\linewidth}{!}{
    \begin{tabular}{l|ccc|ccc}
        \toprule
        \multirow{2}{*}{Method} & \multicolumn{3}{c|}{DAVIS} & \multicolumn{3}{c}{ShutterStock}  \\
         & PSNR$\uparrow$ & SSIM $\uparrow$ &  LPIPS$\downarrow$ &  PSNR$\uparrow$ & SSIM$\uparrow$ &  LPIPS$\downarrow$ \\

        \midrule
        RIFE \cite{huang2022real} & 17.31 & 0.5195 & 0.2512 & 20.44 & 0.7210 & 0.1533 \\
        FILM \cite{reda2022film} & 17.00 & 0.5011 & 0.2363 & 20.90 & 0.7453 & 0.1246 \\
        Ours & 22.15 & 0.6332 & 0.1483 & 24.19 & 0.7616 & 0.1097 \\
        VQGAN (ground truth) & 25.66 & 0.7429 & 0.0784 & 27.81 & 0.8327 & 0.0561 \\

    \bottomrule
    \end{tabular}
    }
    \caption{\small{Quantitative comparisons on video frame interpolation with original keyframes. } }
    \label{tab:rgb_recon}
\end{table}

\vspace{-0.5em}
\paragraph{Number of of keyframes}
Although our model is trained with frame interpolation by default, MaskINT can seamlessly generalize to an arbitrary number of keyframes without finetuning.
We assess the impact of varying the quantity of keyframes on the generation performance. 
As shown in the left part of Table \ref{tab:ablation}, with an increase in the number of keyframes, the model exhibits an improvement in performance. Generally, with more information, performing frame interpolation is easier. 
However, simultaneously editing more frames requires longer time due to the global attention among them.

\vspace{-0.5em}
\paragraph{Decoding steps}
We also explore the number of decoding steps $K$ for the masked generative transformers in the second stage. 
The right part of \cref{tab:ablation} shows that more decoding steps can bring slight improvement on the temporal consistency, but requires more time. Considering the trade-off between performance and efficiency, we chose $K=32$ steps by default in all experiments.

\begin{table}[!ht]

    \centering
    \begin{minipage}{0.45\linewidth}
        \centering
    \resizebox{1.0\linewidth}{!}{
    \begin{tabular}{c|c|c|c}
        \toprule
        \# keyframes  & T.C. & P.C. &  Time \\
        \midrule
        1 & 0.9690 & 0.2984 & 19s \\
        2 & 0.9714 & 0.3038 & 22s \\
        3 &  0.9721 & 0.3051 & 26s \\ 
        4 &  0.9728 & 0.3069 & 29s \\
        6 &  0.9737 & 0.3035 & 35s \\
    \bottomrule
    \end{tabular}
    }  
    \end{minipage}
    \begin{minipage}{0.52\linewidth}
    
    \centering
    \resizebox{1.0\linewidth}{!}{
    \begin{tabular}{c|c|c|c}
        \toprule
        \#decoding step $K$  & T.C. & P.C. &  Time \\
        \midrule
        16 & 0.9691 & 0.3038 & 15s \\
        32 & 0.9714 & 0.3038 & 22s \\
        64 &  0.9719 & 0.3040 & 33s \\
        128 & 0.9720 & 0.3041 & 62s \\
    \bottomrule
    \end{tabular}
    }  
    \end{minipage}
    \caption{ \small{Ablation study on the number of keyframes and the number of decoding steps $K$. ``T.C." stands for ``temporal consistency" and ``P.C." stands for ``prompt consistency".} }
    \label{tab:ablation}
\end{table}



\vspace{-0.9em}
\section{Limitation and Future Work}
\vspace{-0.5em}
One limitation of MaskINT is that it can only perform structure-preserving video editing, such as altering style or appearance. Thus, it cannot handle edits that require structural changes, for example, change a dog to a horse, a limitation shared with TokenFlow \cite{geyer2023tokenflow}. 
It also requires that no new objects should appear in the intermediate frames. 
Besides, the performance of MaskINT highly relies on the tedious image-editing model and structure detector. When these models fail, the structure-aware interpolation becomes meaningless, resulting in artifacts. 
%
In the future, given that our approach disentangles video editing into two distinct stages, we intend to explore the integration of token-based methods, like Muse \cite{chang2023muse}, for image editing. This endeavor aims to further enhance the efficiency of the initial stage.

\section{Conclusion}
\vspace{-0.5em}
We propose MaskINT towards consistent and efficient video editing with prompt. 
MaskINT disentangles this task into keyframes joint editing with T2I model and structure-aware frame interpolation with non-autoregressive masked transformers. 
Experimental results demonstrate that MaskINT achieves comparable performance with pure diffusion-based methods while significantly accelerate the inference up to 7x. 
We believe our work demonstrates the substantial promise of non-autoregressive generative transformers within the realm of video editing.

{
    \small
    \bibliographystyle{ieeenat_fullname}
    \bibliography{main}

\begin{thebibliography}{63}
\providecommand{\natexlab}[1]{#1}
\providecommand{\url}[1]{\texttt{#1}}
\expandafter\ifx\csname urlstyle\endcsname\relax
  \providecommand{\doi}[1]{doi: #1}\else
  \providecommand{\doi}{doi: \begingroup \urlstyle{rm}\Url}\fi

\bibitem[SSd(2023)]{SSdataset}
Stock footage video, royalty-free hd, 4k video clips.
\newblock 2023.

\bibitem[Bain et~al.(2021)Bain, Nagrani, Varol, and Zisserman]{bain2021frozen}
Max Bain, Arsha Nagrani, G{\"u}l Varol, and Andrew Zisserman.
\newblock Frozen in time: A joint video and image encoder for end-to-end retrieval.
\newblock In \emph{ICCV}, 2021.

\bibitem[Bashkirova et~al.(2023)Bashkirova, Lezama, Sohn, Saenko, and Essa]{bashkirova2023masksketch}
Dina Bashkirova, Jos{\'e} Lezama, Kihyuk Sohn, Kate Saenko, and Irfan Essa.
\newblock Masksketch: Unpaired structure-guided masked image generation.
\newblock In \emph{CVPR}, pages 1879--1889, 2023.

\bibitem[Bertasius et~al.(2021)Bertasius, Wang, and Torresani]{bertasius2021space}
Gedas Bertasius, Heng Wang, and Lorenzo Torresani.
\newblock Is space-time attention all you need for video understanding?
\newblock In \emph{ICML}, 2021.

\bibitem[Blattmann et~al.(2023)Blattmann, Dockhorn, Kulal, Mendelevitch, Kilian, Lorenz, Levi, English, Voleti, Letts, et~al.]{blattmann2023stable}
Andreas Blattmann, Tim Dockhorn, Sumith Kulal, Daniel Mendelevitch, Maciej Kilian, Dominik Lorenz, Yam Levi, Zion English, Vikram Voleti, Adam Letts, et~al.
\newblock Stable video diffusion: Scaling latent video diffusion models to large datasets.
\newblock \emph{arXiv preprint arXiv:2311.15127}, 2023.

\bibitem[Brooks et~al.(2023)Brooks, Holynski, and Efros]{brooks2023instructpix2pix}
Tim Brooks, Aleksander Holynski, and Alexei~A Efros.
\newblock Instructpix2pix: Learning to follow image editing instructions.
\newblock In \emph{CVPR}, pages 18392--18402, 2023.

\bibitem[Brown et~al.(2020)Brown, Mann, Ryder, Subbiah, Kaplan, Dhariwal, Neelakantan, Shyam, Sastry, Askell, et~al.]{brown2020language}
Tom Brown, Benjamin Mann, Nick Ryder, Melanie Subbiah, Jared~D Kaplan, Prafulla Dhariwal, Arvind Neelakantan, Pranav Shyam, Girish Sastry, Amanda Askell, et~al.
\newblock Language models are few-shot learners.
\newblock \emph{NeurIPS}, 33:\penalty0 1877--1901, 2020.

\bibitem[Ceylan et~al.(2023)Ceylan, Huang, and Mitra]{ceylan2023pix2video}
Duygu Ceylan, Chun-Hao~P Huang, and Niloy~J Mitra.
\newblock Pix2video: Video editing using image diffusion.
\newblock In \emph{ICCV}, pages 23206--23217, 2023.

\bibitem[Chai et~al.(2023)Chai, Guo, Wang, and Lu]{chai2023stablevideo}
Wenhao Chai, Xun Guo, Gaoang Wang, and Yan Lu.
\newblock Stablevideo: Text-driven consistency-aware diffusion video editing.
\newblock In \emph{Proceedings of the IEEE/CVF International Conference on Computer Vision}, pages 23040--23050, 2023.

\bibitem[Chang et~al.(2022)Chang, Zhang, Jiang, Liu, and Freeman]{chang2022maskgit}
Huiwen Chang, Han Zhang, Lu Jiang, Ce Liu, and William~T Freeman.
\newblock Maskgit: Masked generative image transformer.
\newblock In \emph{CVPR}, pages 11315--11325, 2022.

\bibitem[Chang et~al.(2023)Chang, Zhang, Barber, Maschinot, Lezama, Jiang, Yang, Murphy, Freeman, Rubinstein, et~al.]{chang2023muse}
Huiwen Chang, Han Zhang, Jarred Barber, AJ Maschinot, Jose Lezama, Lu Jiang, Ming-Hsuan Yang, Kevin Murphy, William~T Freeman, Michael Rubinstein, et~al.
\newblock Muse: Text-to-image generation via masked generative transformers.
\newblock \emph{ICML}, 2023.

\bibitem[Couairon et~al.(2023)Couairon, Verbeek, Schwenk, and Cord]{couairon2022diffedit}
Guillaume Couairon, Jakob Verbeek, Holger Schwenk, and Matthieu Cord.
\newblock Diffedit: Diffusion-based semantic image editing with mask guidance.
\newblock \emph{ICLR}, 2023.

\bibitem[Dai et~al.(2023)Dai, Hou, Ma, Tsai, Wang, Wang, Zhang, Vandenhende, Wang, Dubey, et~al.]{dai2023emu}
Xiaoliang Dai, Ji Hou, Chih-Yao Ma, Sam Tsai, Jialiang Wang, Rui Wang, Peizhao Zhang, Simon Vandenhende, Xiaofang Wang, Abhimanyu Dubey, et~al.
\newblock Emu: Enhancing image generation models using photogenic needles in a haystack.
\newblock \emph{arXiv preprint arXiv:2309.15807}, 2023.

\bibitem[Devlin et~al.(2019)Devlin, Chang, Lee, and Toutanova]{devlin2018bert}
Jacob Devlin, Ming-Wei Chang, Kenton Lee, and Kristina Toutanova.
\newblock Bert: Pre-training of deep bidirectional transformers for language understanding.
\newblock \emph{NAACL}, 2019.

\bibitem[Dubey et~al.(2023)Dubey, Radenovic, Mahajan, and Ramanathan]{dubeyretinavq}
Abhimanyu Dubey, Filip Radenovic, Dhruv Mahajan, and Vignesh Ramanathan.
\newblock Retina vq.
\newblock 2023.

\bibitem[Esser et~al.(2021)Esser, Rombach, and Ommer]{esser2021taming}
Patrick Esser, Robin Rombach, and Bjorn Ommer.
\newblock Taming transformers for high-resolution image synthesis.
\newblock In \emph{CVPR}, pages 12873--12883, 2021.

\bibitem[Esser et~al.(2023)Esser, Chiu, Atighehchian, Granskog, and Germanidis]{esser2023structure}
Patrick Esser, Johnathan Chiu, Parmida Atighehchian, Jonathan Granskog, and Anastasis Germanidis.
\newblock Structure and content-guided video synthesis with diffusion models.
\newblock \emph{ICCV}, 2023.

\bibitem[Ge et~al.(2022)Ge, Hayes, Yang, Yin, Pang, Jacobs, Huang, and Parikh]{ge2022long}
Songwei Ge, Thomas Hayes, Harry Yang, Xi Yin, Guan Pang, David Jacobs, Jia-Bin Huang, and Devi Parikh.
\newblock Long video generation with time-agnostic vqgan and time-sensitive transformer.
\newblock In \emph{ECCV}, pages 102--118. Springer, 2022.

\bibitem[Geyer et~al.(2024)Geyer, Bar-Tal, Bagon, and Dekel]{geyer2023tokenflow}
Michal Geyer, Omer Bar-Tal, Shai Bagon, and Tali Dekel.
\newblock Tokenflow: Consistent diffusion features for consistent video editing.
\newblock \emph{ICLR}, 2024.

\bibitem[Gupta et~al.(2023)Gupta, Tian, Zhang, Wu, Mart{\'\i}n-Mart{\'\i}n, and Fei-Fei]{gupta2023maskvit}
Agrim Gupta, Stephen Tian, Yunzhi Zhang, Jiajun Wu, Roberto Mart{\'\i}n-Mart{\'\i}n, and Li Fei-Fei.
\newblock Maskvit: Masked visual pre-training for video prediction.
\newblock In \emph{ICLR}, 2023.

\bibitem[Hertz et~al.(2022)Hertz, Mokady, Tenenbaum, Aberman, Pritch, and Cohen-Or]{hertz2022prompt}
Amir Hertz, Ron Mokady, Jay Tenenbaum, Kfir Aberman, Yael Pritch, and Daniel Cohen-Or.
\newblock Prompt-to-prompt image editing with cross attention control.
\newblock \emph{arXiv preprint arXiv:2208.01626}, 2022.

\bibitem[Ho et~al.(2020)Ho, Jain, and Abbeel]{ho2020denoising}
Jonathan Ho, Ajay Jain, and Pieter Abbeel.
\newblock Denoising diffusion probabilistic models.
\newblock \emph{NeurIPS}, 33:\penalty0 6840--6851, 2020.

\bibitem[Ho et~al.(2022)Ho, Chan, Saharia, Whang, Gao, Gritsenko, Kingma, Poole, Norouzi, Fleet, et~al.]{ho2022imagen}
Jonathan Ho, William Chan, Chitwan Saharia, Jay Whang, Ruiqi Gao, Alexey Gritsenko, Diederik~P Kingma, Ben Poole, Mohammad Norouzi, David~J Fleet, et~al.
\newblock Imagen video: High definition video generation with diffusion models.
\newblock \emph{arXiv preprint arXiv:2210.02303}, 2022.

\bibitem[Hong et~al.(2023)Hong, Ding, Zheng, Liu, and Tang]{hong2022cogvideo}
Wenyi Hong, Ming Ding, Wendi Zheng, Xinghan Liu, and Jie Tang.
\newblock Cogvideo: Large-scale pretraining for text-to-video generation via transformers.
\newblock \emph{ICLR}, 2023.

\bibitem[Huang et~al.(2023)Huang, Chen, Liu, Shen, Zhao, and Zhou]{huang2023composer}
Lianghua Huang, Di Chen, Yu Liu, Yujun Shen, Deli Zhao, and Jingren Zhou.
\newblock Composer: Creative and controllable image synthesis with composable conditions.
\newblock \emph{arXiv preprint arXiv:2302.09778}, 2023.

\bibitem[Huang et~al.(2022)Huang, Zhang, Heng, Shi, and Zhou]{huang2022real}
Zhewei Huang, Tianyuan Zhang, Wen Heng, Boxin Shi, and Shuchang Zhou.
\newblock Real-time intermediate flow estimation for video frame interpolation.
\newblock In \emph{ECCV}, pages 624--642, 2022.

\bibitem[Jiang et~al.(2018)Jiang, Sun, Jampani, Yang, Learned-Miller, and Kautz]{jiang2018super}
Huaizu Jiang, Deqing Sun, Varun Jampani, Ming-Hsuan Yang, Erik Learned-Miller, and Jan Kautz.
\newblock Super slomo: High quality estimation of multiple intermediate frames for video interpolation.
\newblock In \emph{CVPR}, pages 9000--9008, 2018.

\bibitem[Kawar et~al.(2023)Kawar, Zada, Lang, Tov, Chang, Dekel, Mosseri, and Irani]{kawar2023imagic}
Bahjat Kawar, Shiran Zada, Oran Lang, Omer Tov, Huiwen Chang, Tali Dekel, Inbar Mosseri, and Michal Irani.
\newblock Imagic: Text-based real image editing with diffusion models.
\newblock In \emph{CVPR}, pages 6007--6017, 2023.

\bibitem[Khachatryan et~al.(2023)Khachatryan, Movsisyan, Tadevosyan, Henschel, Wang, Navasardyan, and Shi]{khachatryan2023text2video}
Levon Khachatryan, Andranik Movsisyan, Vahram Tadevosyan, Roberto Henschel, Zhangyang Wang, Shant Navasardyan, and Humphrey Shi.
\newblock Text2video-zero: Text-to-image diffusion models are zero-shot video generators.
\newblock \emph{ICCV}, 2023.

\bibitem[Lai et~al.(2018)Lai, Huang, Wang, Shechtman, Yumer, and Yang]{lai2018learning}
Wei-Sheng Lai, Jia-Bin Huang, Oliver Wang, Eli Shechtman, Ersin Yumer, and Ming-Hsuan Yang.
\newblock Learning blind video temporal consistency.
\newblock In \emph{ECCV}, 2018.

\bibitem[Le~Moing et~al.(2021)Le~Moing, Ponce, and Schmid]{le2021ccvs}
Guillaume Le~Moing, Jean Ponce, and Cordelia Schmid.
\newblock Ccvs: context-aware controllable video synthesis.
\newblock \emph{NeurIPS}, 34:\penalty0 14042--14055, 2021.

\bibitem[Liew et~al.(2023)Liew, Yan, Zhang, Xu, and Feng]{liew2023magicedit}
Jun~Hao Liew, Hanshu Yan, Jianfeng Zhang, Zhongcong Xu, and Jiashi Feng.
\newblock Magicedit: High-fidelity and temporally coherent video editing.
\newblock \emph{arXiv preprint arXiv:2308.14749}, 2023.

\bibitem[Loshchilov and Hutter(2019)]{loshchilov2018decoupled}
Ilya Loshchilov and Frank Hutter.
\newblock Decoupled weight decay regularization.
\newblock In \emph{ICLR}, 2019.

\bibitem[Lu et~al.(2022)Lu, Wu, Lin, Lu, and Jia]{lu2022video}
Liying Lu, Ruizheng Wu, Huaijia Lin, Jiangbo Lu, and Jiaya Jia.
\newblock Video frame interpolation with transformer.
\newblock In \emph{CVPR}, pages 3532--3542, 2022.

\bibitem[Meng et~al.(2022)Meng, He, Song, Song, Wu, Zhu, and Ermon]{meng2021sdedit}
Chenlin Meng, Yutong He, Yang Song, Jiaming Song, Jiajun Wu, Jun-Yan Zhu, and Stefano Ermon.
\newblock Sdedit: Guided image synthesis and editing with stochastic differential equations.
\newblock \emph{ICLR}, 2022.

\bibitem[Mou et~al.(2023)Mou, Wang, Xie, Zhang, Qi, Shan, and Qie]{mou2023t2i}
Chong Mou, Xintao Wang, Liangbin Xie, Jian Zhang, Zhongang Qi, Ying Shan, and Xiaohu Qie.
\newblock T2i-adapter: Learning adapters to dig out more controllable ability for text-to-image diffusion models.
\newblock \emph{arXiv preprint arXiv:2302.08453}, 2023.

\bibitem[Niklaus et~al.(2017)Niklaus, Mai, and Liu]{niklaus2017video}
Simon Niklaus, Long Mai, and Feng Liu.
\newblock Video frame interpolation via adaptive separable convolution.
\newblock In \emph{ICCV}, pages 261--270, 2017.

\bibitem[Parmar et~al.(2023)Parmar, Kumar~Singh, Zhang, Li, Lu, and Zhu]{parmar2023zero}
Gaurav Parmar, Krishna Kumar~Singh, Richard Zhang, Yijun Li, Jingwan Lu, and Jun-Yan Zhu.
\newblock Zero-shot image-to-image translation.
\newblock In \emph{ACM SIGGRAPH 2023 Conference Proceedings}, pages 1--11, 2023.

\bibitem[Peebles and Xie(2023)]{peebles2023scalable}
William Peebles and Saining Xie.
\newblock Scalable diffusion models with transformers.
\newblock In \emph{ICCV}, pages 4195--4205, 2023.

\bibitem[Pont-Tuset et~al.(2017)Pont-Tuset, Perazzi, Caelles, Arbel\'aez, Sorkine-Hornung, and {Van Gool}]{Pont-Tuset_arXiv_2017}
Jordi Pont-Tuset, Federico Perazzi, Sergi Caelles, Pablo Arbel\'aez, Alexander Sorkine-Hornung, and Luc {Van Gool}.
\newblock The 2017 davis challenge on video object segmentation.
\newblock \emph{arXiv:1704.00675}, 2017.

\bibitem[Qi et~al.(2023)Qi, Cun, Zhang, Lei, Wang, Shan, and Chen]{qi2023fatezero}
Chenyang Qi, Xiaodong Cun, Yong Zhang, Chenyang Lei, Xintao Wang, Ying Shan, and Qifeng Chen.
\newblock Fatezero: Fusing attentions for zero-shot text-based video editing.
\newblock \emph{ICCV}, 2023.

\bibitem[Radford et~al.(2021)Radford, Kim, Hallacy, Ramesh, Goh, Agarwal, Sastry, Askell, Mishkin, Clark, et~al.]{radford2021learning}
Alec Radford, Jong~Wook Kim, Chris Hallacy, Aditya Ramesh, Gabriel Goh, Sandhini Agarwal, Girish Sastry, Amanda Askell, Pamela Mishkin, Jack Clark, et~al.
\newblock Learning transferable visual models from natural language supervision.
\newblock In \emph{ICML}, pages 8748--8763, 2021.

\bibitem[Reda et~al.(2022)Reda, Kontkanen, Tabellion, Sun, Pantofaru, and Curless]{reda2022film}
Fitsum Reda, Janne Kontkanen, Eric Tabellion, Deqing Sun, Caroline Pantofaru, and Brian Curless.
\newblock Film: Frame interpolation for large motion.
\newblock In \emph{ECCV}, 2022.

\bibitem[Rombach et~al.(2022)Rombach, Blattmann, Lorenz, Esser, and Ommer]{rombach2022high}
Robin Rombach, Andreas Blattmann, Dominik Lorenz, Patrick Esser, and Bj{\"o}rn Ommer.
\newblock High-resolution image synthesis with latent diffusion models.
\newblock In \emph{CVPR}, pages 10684--10695, 2022.

\bibitem[Saharia et~al.(2022)Saharia, Chan, Saxena, Li, Whang, Denton, Ghasemipour, Gontijo~Lopes, Karagol~Ayan, Salimans, et~al.]{saharia2022photorealistic}
Chitwan Saharia, William Chan, Saurabh Saxena, Lala Li, Jay Whang, Emily~L Denton, Kamyar Ghasemipour, Raphael Gontijo~Lopes, Burcu Karagol~Ayan, Tim Salimans, et~al.
\newblock Photorealistic text-to-image diffusion models with deep language understanding.
\newblock \emph{NeurIPS}, 35:\penalty0 36479--36494, 2022.

\bibitem[Sim et~al.(2021)Sim, Oh, and Kim]{sim2021xvfi}
Hyeonjun Sim, Jihyong Oh, and Munchurl Kim.
\newblock Xvfi: extreme video frame interpolation.
\newblock In \emph{ICCV}, pages 14489--14498, 2021.

\bibitem[Singer et~al.(2023)Singer, Polyak, Hayes, Yin, An, Zhang, Hu, Yang, Ashual, Gafni, et~al.]{singer2022make}
Uriel Singer, Adam Polyak, Thomas Hayes, Xi Yin, Jie An, Songyang Zhang, Qiyuan Hu, Harry Yang, Oron Ashual, Oran Gafni, et~al.
\newblock Make-a-video: Text-to-video generation without text-video data.
\newblock \emph{ICLR}, 2023.

\bibitem[Sohn et~al.(2023)Sohn, Ruiz, Lee, Chin, Blok, Chang, Barber, Jiang, Entis, Li, et~al.]{sohn2023styledrop}
Kihyuk Sohn, Nataniel Ruiz, Kimin Lee, Daniel~Castro Chin, Irina Blok, Huiwen Chang, Jarred Barber, Lu Jiang, Glenn Entis, Yuanzhen Li, et~al.
\newblock Styledrop: Text-to-image generation in any style.
\newblock \emph{arXiv preprint arXiv:2306.00983}, 2023.

\bibitem[Song et~al.(2021)Song, Meng, and Ermon]{song2020denoising}
Jiaming Song, Chenlin Meng, and Stefano Ermon.
\newblock Denoising diffusion implicit models.
\newblock \emph{ICLR}, 2021.

\bibitem[Teed and Deng(2020)]{teed2020raft}
Zachary Teed and Jia Deng.
\newblock Raft: Recurrent all-pairs field transforms for optical flow.
\newblock In \emph{ECCV}, 2020.

\bibitem[Tumanyan et~al.(2023)Tumanyan, Geyer, Bagon, and Dekel]{tumanyan2023plug}
Narek Tumanyan, Michal Geyer, Shai Bagon, and Tali Dekel.
\newblock Plug-and-play diffusion features for text-driven image-to-image translation.
\newblock In \emph{CVPR}, pages 1921--1930, 2023.

\bibitem[Tzaban et~al.(2022)Tzaban, Mokady, Gal, Bermano, and Cohen-Or]{tzaban2022stitch}
Rotem Tzaban, Ron Mokady, Rinon Gal, Amit Bermano, and Daniel Cohen-Or.
\newblock Stitch it in time: Gan-based facial editing of real videos.
\newblock In \emph{SIGGRAPH Asia}, 2022.

\bibitem[Vaswani et~al.(2017)Vaswani, Shazeer, Parmar, Uszkoreit, Jones, Gomez, Kaiser, and Polosukhin]{vaswani2017attention}
Ashish Vaswani, Noam Shazeer, Niki Parmar, Jakob Uszkoreit, Llion Jones, Aidan~N Gomez, {\L}ukasz Kaiser, and Illia Polosukhin.
\newblock Attention is all you need.
\newblock \emph{NeurIPS}, 30, 2017.

\bibitem[Villegas et~al.(2023)Villegas, Babaeizadeh, Kindermans, Moraldo, Zhang, Saffar, Castro, Kunze, and Erhan]{villegas2023phenaki}
Ruben Villegas, Mohammad Babaeizadeh, Pieter-Jan Kindermans, Hernan Moraldo, Han Zhang, Mohammad~Taghi Saffar, Santiago Castro, Julius Kunze, and Dumitru Erhan.
\newblock Phenaki: Variable length video generation from open domain textual descriptions.
\newblock In \emph{ICLR}, 2023.

\bibitem[Wang et~al.(2023)Wang, Yuan, Zhang, Chen, Wang, Zhang, Shen, Zhao, and Zhou]{wang2023videocomposer}
Xiang Wang, Hangjie Yuan, Shiwei Zhang, Dayou Chen, Jiuniu Wang, Yingya Zhang, Yujun Shen, Deli Zhao, and Jingren Zhou.
\newblock Videocomposer: Compositional video synthesis with motion controllability.
\newblock \emph{NeurIPS}, 2023.

\bibitem[Wu et~al.(2023)Wu, Ge, Wang, Lei, Gu, Hsu, Shan, Qie, and Shou]{wu2022tune}
Jay~Zhangjie Wu, Yixiao Ge, Xintao Wang, Weixian Lei, Yuchao Gu, Wynne Hsu, Ying Shan, Xiaohu Qie, and Mike~Zheng Shou.
\newblock Tune-a-video: One-shot tuning of image diffusion models for text-to-video generation.
\newblock \emph{ICCV}, 2023.

\bibitem[Xie and Tu(2015)]{xie2015holistically}
Saining Xie and Zhuowen Tu.
\newblock Holistically-nested edge detection.
\newblock In \emph{ICCV}, pages 1395--1403, 2015.

\bibitem[Yang et~al.(2023)Yang, Zhou, Liu, and Loy]{yang2023rerender}
Shuai Yang, Yifan Zhou, Ziwei Liu, and Chen~Change Loy.
\newblock Rerender a video: Zero-shot text-guided video-to-video translation.
\newblock \emph{ACM SIGGRAPH Asia}, 2023.

\bibitem[Yu et~al.(2022)Yu, Li, Koh, Zhang, Pang, Qin, Ku, Xu, Baldridge, and Wu]{yu2022vectorquantized}
Jiahui Yu, Xin Li, Jing~Yu Koh, Han Zhang, Ruoming Pang, James Qin, Alexander Ku, Yuanzhong Xu, Jason Baldridge, and Yonghui Wu.
\newblock Vector-quantized image modeling with improved {VQGAN}.
\newblock In \emph{ICLR}, 2022.

\bibitem[Yu et~al.(2023)Yu, Cheng, Sohn, Lezama, Zhang, Chang, Hauptmann, Yang, Hao, Essa, et~al.]{yu2023magvit}
Lijun Yu, Yong Cheng, Kihyuk Sohn, Jos{\'e} Lezama, Han Zhang, Huiwen Chang, Alexander~G Hauptmann, Ming-Hsuan Yang, Yuan Hao, Irfan Essa, et~al.
\newblock Magvit: Masked generative video transformer.
\newblock In \emph{CVPR}, pages 10459--10469, 2023.

\bibitem[Zhang and Agrawala(2023)]{zhang2023adding}
Lvmin Zhang and Maneesh Agrawala.
\newblock Adding conditional control to text-to-image diffusion models.
\newblock \emph{ICCV}, 2023.

\bibitem[Zhang et~al.(2024)Zhang, Wei, Jiang, Zhang, Zuo, and Tian]{zhang2023controlvideo}
Yabo Zhang, Yuxiang Wei, Dongsheng Jiang, Xiaopeng Zhang, Wangmeng Zuo, and Qi Tian.
\newblock Controlvideo: Training-free controllable text-to-video generation.
\newblock \emph{ICLR}, 2024.

\bibitem[Zhang et~al.(2023)Zhang, Han, Ghosh, Metaxas, and Ren]{zhang2023sine}
Zhixing Zhang, Ligong Han, Arnab Ghosh, Dimitris~N Metaxas, and Jian Ren.
\newblock Sine: Single image editing with text-to-image diffusion models.
\newblock In \emph{CVPR}, pages 6027--6037, 2023.

\end{thebibliography}
}


\clearpage
\setcounter{page}{1}
\setcounter{section}{0}
\setcounter{table}{0}
\setcounter{figure}{0}

\maketitlesupplementary

\section{Dataset Details}
ShutterStock \cite{SSdataset} is a commercial dataset with over 30 million paired text-video examples. Our model is trained using a subset of 100k videos. 
The WebVid dataset \cite{bain2021frozen} offers an accessible version of ShutterStock, but they include a central watermark due to copyright constraints. 

\section{Diverse Structural Conditions}
Although we by default utilize HED edge map as our structure condition for both stages, our method can also employ other structural controls in both stages due to the disentanglement. 
In this study, we explore utilizing ControlNet with depth map to perform key frame editing, and using the depth map as the guidance to perform structure-aware frame interpolation. Additionally, we explore various combinations of these approaches. 
As summarized in Table \ref{tab:combination}, all of these combinations achieve the same level of performance. The performance of prompt consistency (P.C.) is determined by the specific key frame editing methods employed. 
For the second stage, depth control typically offers greater flexibility than HED edge control for frame interpolation. This could be the potential reason for the slightly worse performance in temporal consistency with edge-based key frame editing.

\begin{table}[!ht]
    \centering
    \resizebox{0.8\linewidth}{!}{
    \begin{tabular}{ll|cc}
        \toprule
         Stage1 & Stage2 & T.C. & P.C \\
         \midrule
         HED edge & HED edge & 0.9714 & 0.3038 \\
         depth map & HED edge & 0.9713 & 0.3159 \\
         HED edge & depth map & 0.9683 & 0.3035 \\
         depth map & depth map & 0.9719 & 0.3171 \\
    \bottomrule
    \end{tabular}
    }
    \caption{\small{Quantitative comparisons of the combination of varied structural conditions in each stage on ShutterStock. ``T.C." stands for ``temporal consistency", and ``P.C." stands for ``prompt consistency".} }
    \label{tab:combination}
\end{table}

\section{Additional Examples for Comparisons}
\begin{figure*}[!ht]
    \centering
    \includegraphics[width=1.0\linewidth]{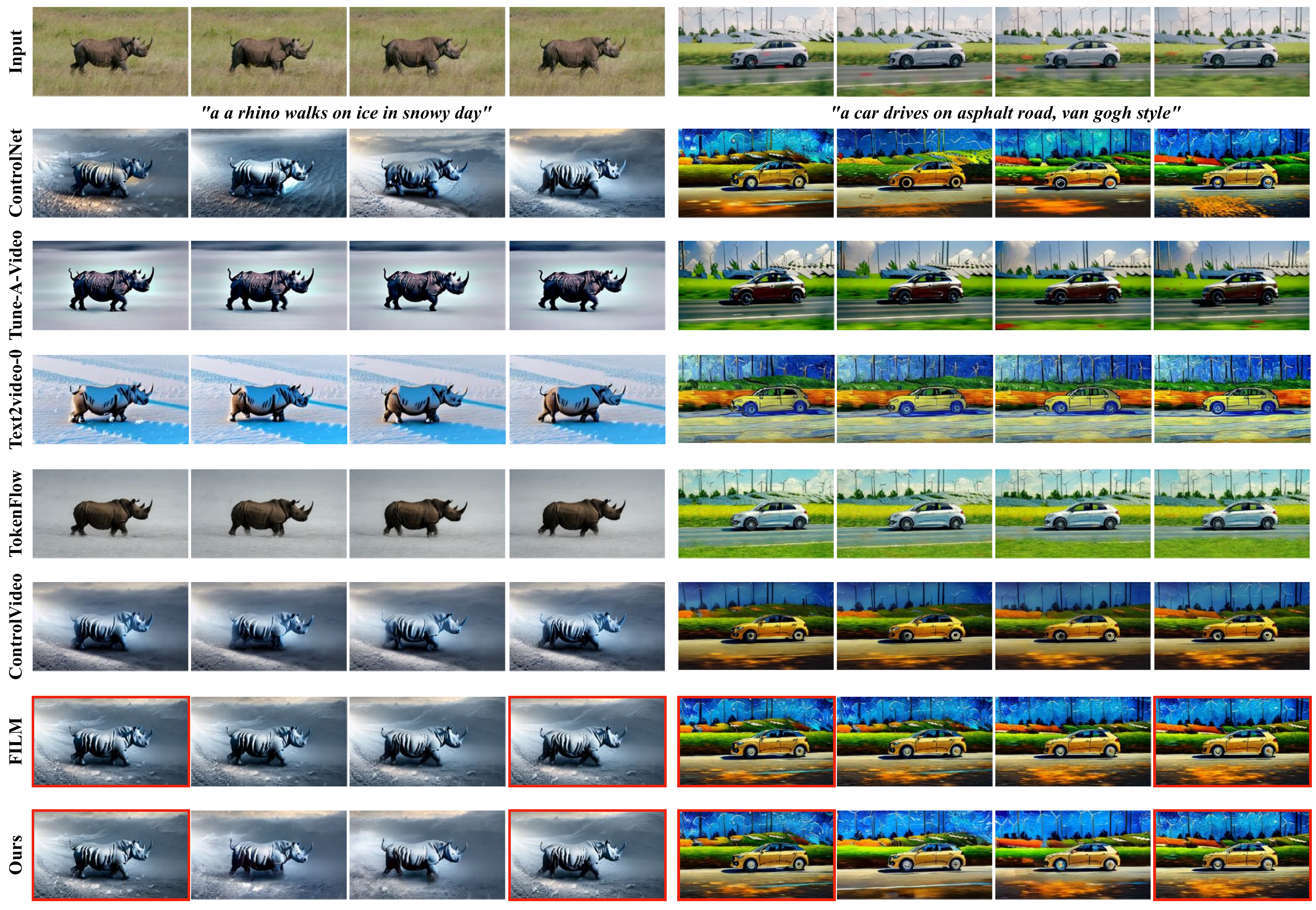}
    \caption{ \small{Additional Qualitative comparisons with diffusion-based methods. Frames with red bounding box are jointly edited keyeframes. } }
    \label{fig:cmp_sup1}
\end{figure*}

We add more comparisons with diffusion methods in Fig. \ref{fig:cmp_sup1}. Our methods can maintain the temporal consistency among variant examples like other pure-diffusion methods.

\section{Additional Editing Examples}
\begin{figure*}[!ht]
    \centering
    \includegraphics[width=1.0\linewidth]{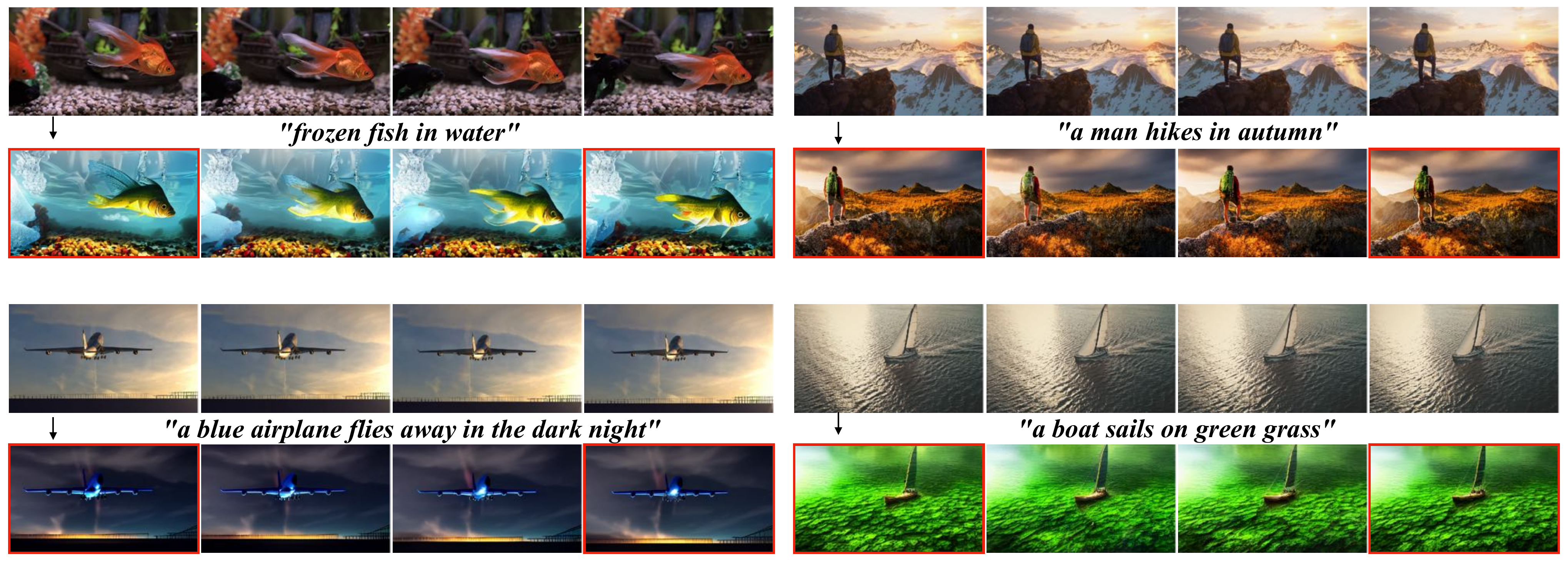}
    \caption{ \small{Additional Editing examples with MaskINT. Frames with red bounding box are jointly edited keyeframes. } }
    \label{fig:more_samples}
\end{figure*}

We present more video editing examples in Figure \ref{fig:more_samples} to show the generalization of our method.

\section{Example of Failure Cases}
\begin{figure}[ht]
    \centering
    \includegraphics[width=1.0\linewidth]{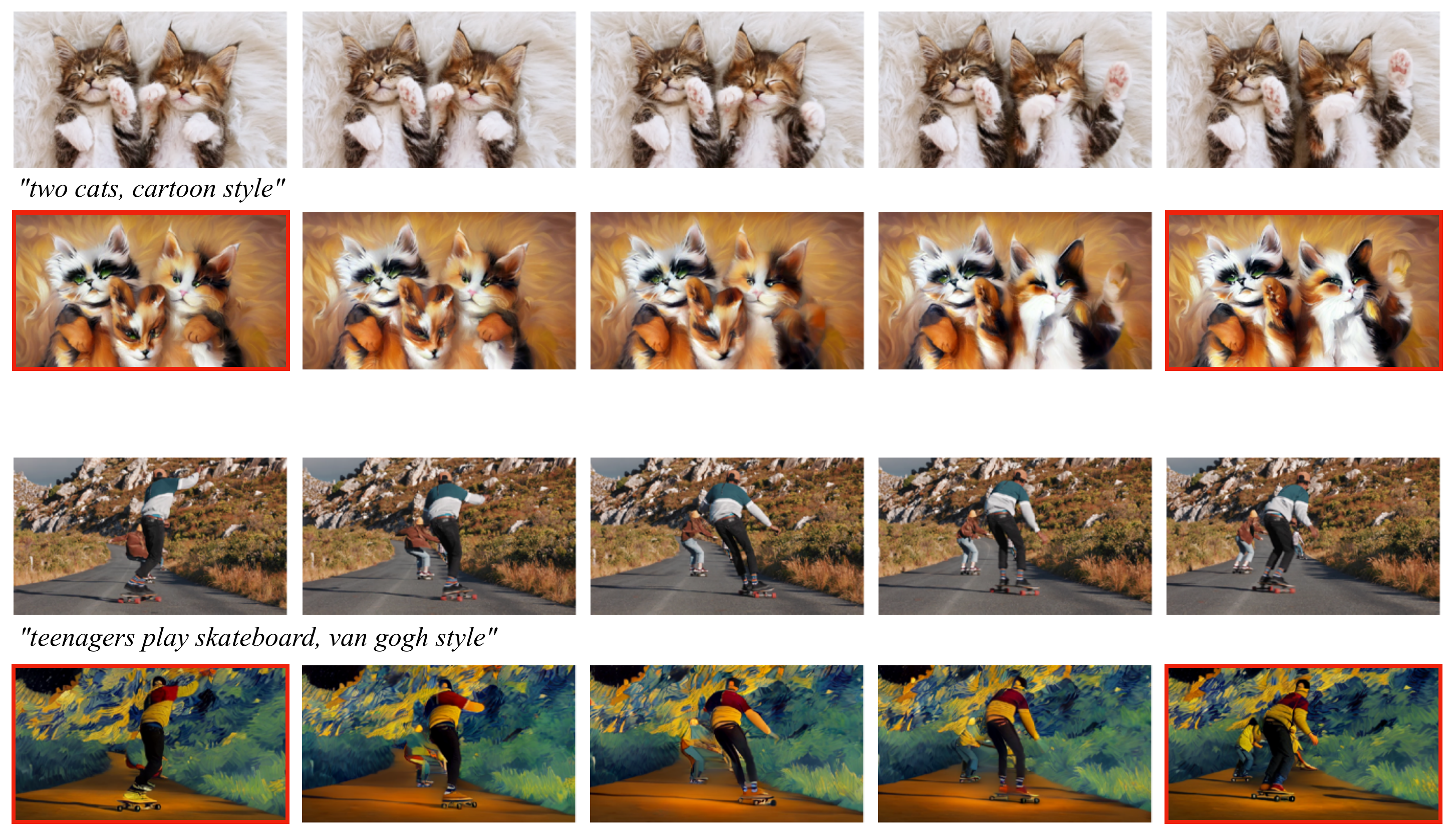}
    \caption{ \small{Examples of failure cases. } }
    \label{fig:fail}
\end{figure}

Since we disentangle the video editing tasks into two separate stage, the final performance of the generated video depends on the key frame editing in the first stage. In certain challenging scenarios, the attention-based key frame editing stage struggles to produce consistent frames, primarily due to the complexity of the scene or the presence of exceptionally large motion. 
In this case, our MaskINT can still interpolate the intermediate frames, albeit with the potential for introducing artifacts. 
Figure \ref{fig:fail} show some failure cases when the first stage fails. 

\end{document}